# A Novel Method for Developing Robotics via Artificial Intelligence and Internet of Things

Aadhityan A
UG Student, TBML College,
Porayar,
Tamil Nadu,
India.

## ABSTRACT
This paper describe about a new methodology for developing and improving the robotics field via artificial intelligence and internet of things. Now a day, we can say Artificial Intelligence take the world into robotics. Almost all industries use robots for lot of works. They are use co-operative robots to make different kind of works. But there was some problem to make robot for multi tasks. So there was a necessary new methodology to made multi tasking robots. It will be done only by artificial intelligence and internet of things.

## 1. INTRODUCTION
Machines as intelligent as humans should be able to do most of the things humans can do. Flexible automation schemes and systems are quickly gaining acceptance and are being increasingly applied in industry. Now a day, they are use robots for manufacturing process. Applications of Artificial Intelligence to medicine have been increasing in recent years. Artificial Intelligence employed in the medical field mainly to perform a rational analysis of the data available from clinical examination (such as Computerized Tomography, Nuclear Magnetic Resonance(NMR), Digital Angiography, etc.) in order to get a reliable diagnosis of the patient's disease and a suitable decision support. Also lot of robots invented to work in home. Robots also invented to play chess, soccer etc. Robots can think well when we implement artificial intelligent concept to it. This paper describes a method to make robots using artificial intelligence.

## 2. SHORT HISTORY OF ROBOTICS
The truck was invented in eighteenth centuries. So we can say making robots using artificial intelligence is old one. The truck is a chess playing robot.

Robotics are based on two enabling technologies: Telemanipulators and the ability of numerical control of machines. Telemanipulators are remotely controlled machines which usually consist of an arm and a gripper. The movements of arm and gripper follow the instructions the human gives through his control device. First telemanipulators have been used to deal with radio-active material. Numeric control allows tocontrol machines very precisely in relation to a given coordinate system. It was first used in 1952 at the MIT and lead to the first programming language for machines (called APT: Automatic Programmed Tools).

The combination of both of these techniques leads to the first programmable telemanipulator.

The first industrial robot using these principles was installed in 1961. These are the robots one knows from industrial facilities like car construction plants.

The development of mobile robots was driven by the desire to automate transportation in production processes and autonomous transport systems. The former lead to driver-less transport systems used on factory floors to move objects to different points in the production process in the late seventies. New forms of mobile robots have been constructed lately like insectoid robots with many legs modeled after examples nature gave us or autonomous robots for underwater usage.

Since a few years wheel-driven robots are commercially marketed and used for services like "Get and Bring" (for example in hospitals). Humanoid robots are being developed since 1975 when Wabot-I was presented in Japan. Also another humanoid robot is "Cog", developed in the MIT-AI-Lab since 1994. Cog has no legs, but it has robotic arms and a head with video cameras for eyes. Honda's humanoid robot became well known in the public when presented back in 1999. Although it is remote controlled by humans it can walk autonomously (on the floor and stairs).

## 3. HISTORY OF ARTIFICIAL INTELLIGENCE
The first work that is now generally recognized as AI was done by Warren McCulloch and Walter Pitts (1943). They drew on three sources: knowledge of the basic physiology and function of neurons in the brain; a formal analysis of propositional logic due to Russell and Whitehead; and Turing's theory of computation. They proposed a model of artificial neurons in which each neuron is characterized as being "on" or "off," with a switch to "on" occurring in response to stimulation by a sufficient number of neighboring neurons. The state of a neuron was conceived of as "factually equivalent to a proposition which proposed its adequate stimulus." They showed, for example, that any computable function could be computed by some network of connected neurons, and that all the logical connectives (and, or, not, etc.) could be implemented by simple net structures. McCulloch and Pitts also suggested that suitably defined networks could learn. Donald Hebb (1949) demonstrated a simple updating rule for modifying the connection strengths between neurons. His rule, now called Hebbian learning, remains an influential model to this day.

AI currently encompasses a huge variety of subfields, ranging from the general (learning and perception) to the specific, such as playing chess, proving mathematical theorems, writing poetry, driving a car on a crowded street, and diagnosing diseases. AI is relevant to any intellectual task; it is truly a universal field.

## 4. DEFINATION TO GENERAL TERMS
### 4.1 Artificial Intelligence
The branch of computer science concerned with making computers behave like humans. The term was coined in 1956 by John McCarthy at the Massachusetts Institute of Technology. Artificial intelligence includes games playing: programming computers to play games such as chess and checkers expert systems,(programming computers to make decisions in real-life tasks (for example, some expert systems help doctors diagnose diseases based on symptoms)) natural language,(programming computers to understand natural human languages) neural networks(Systems that simulate intelligence by attempting to reproduce the types of physical connections that occur in animal brains) and robotics( programming computers to see and hear and react to other sensory stimuli).

### 4.2 Agents
An agent is anything that can be viewed as perceiving its environment through sensor and acting that environment through effectors. A human agent has eyes, ears, nose and other organs for sensors, and hands, legs, mouth, and other body parts for effectors. A robotic agent substitutes the cameras and infrared range finders for the sensors and various motors for the effectors.

### 4.3 Robots
The term "robot" generally connotes some anthropomorphic (human-like) appearance; consider robot "arms" for welding. The tendency to think about robots as having a human-like appearance may stem from the origin of the term "robot." The word "robot" came into the popular consciousness on January 25, 1921, in Prague with the first performance of Karel Capek's play, R.U.R. (Rossum's Universal Robots).

### 4.4 Internet of Things
The next wave in the era of computing field will be outside the realm of the traditional desktop. At present, the definitions of "Internet of Things" are manifold; they vary depending on the context, the effects and the views of the person giving the definition. But we can say, Internet of Things (IoT) paradigm, many of the objects which surround us will be on the network in one form or another. Radio Frequency IDentification (RFID) and sensors network technologies will rise to meet these new challenges, in which information and communication systems are invisibly entrenched(embedded) in the environment around us.

## 5. Method
### 5.1 Analyzing tasks
Analyzing the tasks is very complicated process. The tasks which we want to implement to make a robot can be writing down as a points or table. So we can easily found out how many tasks we want to implement to that robot. It is a general way and it is fairly used to make lot of type of robots in this world.

### 5.2 Concept to make robot
The general concepts to make the robot as a given tasks is the second step in this method. The concepts are involving the sensors which are used to make a robot according to given tasks. Programming the robot also considered as a general concept. So, programming the sensors and other parts (like microprocessor, some chip sets etc.) are also including in this concept. For example, to make a robot which can take a thing. The sensors need to make the robot and programming the dimension of that thing has been taken general concepts to make a robot.

### 5.3 Concept via artificial intelligence
There is needed to make robots which can learn from it own environment. It is very important concept of artificial intelligence. So, programming artificial concepts to each task separately is very important .In this process, we don't consider the hardware pats and sensors. Concepts involving artificial intelligence in each task are found out in this process. These artificial intelligence concepts can be find out directly from the tasks and also by the general concepts to make a robot.

### 5.4 Binding the concepts to make robot
Artificial intelligence concepts and the general concepts in each task are bind together to make a robot which can capable to think and do like a human. In this process, we considered the sensors which are capable to implement our artificial intelligence concept. There is may be some extra sensor or extra requirements needed to implement artificial concepts to robot other than listed in the second step (see 4.2). In this process, we bind the concepts and also we find out the full requirements to make robots which are able to think like a human.

### 5.5 Make a robot
Using all above process, the robot is made. All hardware parts of the robots connected and the programs also written in the sensors and other parts.

### 5.6 Make it to connect with other systems or robots
Finally we can make the robot to connect other system or robots using infrared or blue tooth or some other wireless devices. We can use these wireless devices to connect to mobile or computers. So, the robot can be accessed from mobile, computer or other similar devices.

This method and its process are clearly shown in the fig. 1.

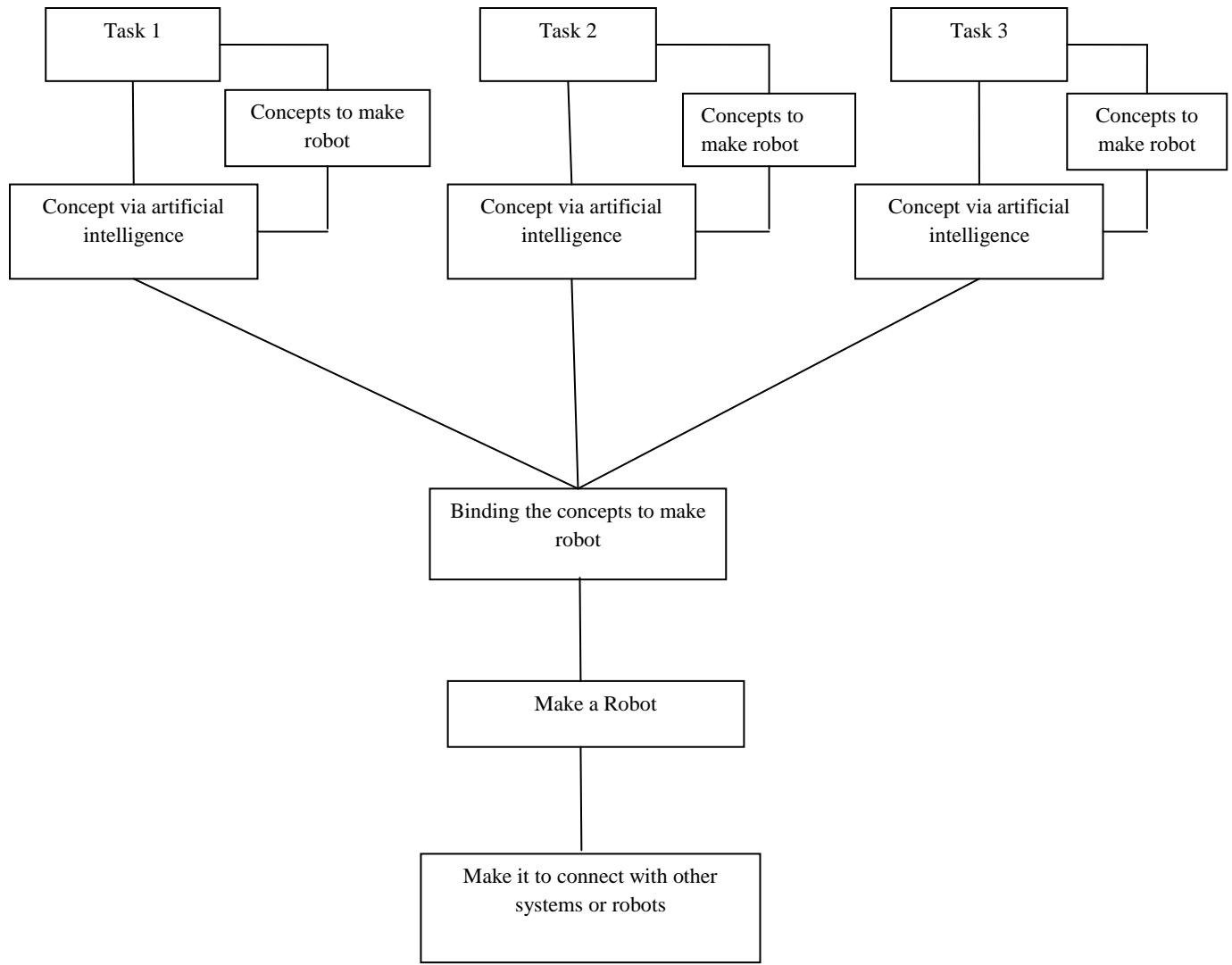

**Fig 1: Method for making robot via artificial intelligence and internet of things.**

## 6. Conclusion
Now a day, robots are highly needed for fast manufacturing. Using this method, we can make robots easily and efficiently.

## 7. ACKNOWLEDGMENTS
I thank to everyone helped in this work. Particularly I would like to thank Dr. Jobe Prabakar Ph.D., who is very helpful to me in entire process. I also want to thank Josuva for his help. I also give thank to S.Arivazhagan and Kalaimani for help to make this paper.

## 8. REFERENCES
[1] Tim Niemueller and Sumedha Widyadharma "Artificial Intelligence – An Introduction to Robotics". Reasoning about naming systems.

[2] Jayavardhana Gubbia, Rajkumar Buyyab,∗, Slaven Marusic a, Marimuthu Palaniswami 2013 "Internet of Things (IoT): A vision, architectural elements, and future directions" Future Generation Computer Systems,Elsevier.

[3] Stuart J. Russell and Peter Norvig 1995. "Artificial Intelligence-A Modern Approach". Prentice-Hall, Inc.

[4] ROBINR. MURPHY, 2000 "Introduction to AI robotics" Massachusetts Institute of technology.

[5] Bruce G. Buchanan, 2005, "A (very) Brief History Of Artificial Intelligence" American Association for Artificial Intelligence.

[6] Elena ALESSANDRI and ALESSANDRO GASPARETTO, RAFAEL VALENCIA GARCIA and RODRIGO MARTINEZ BÉJAR. 2003. "An Application of Artificial Intelligence to Medical Robotics"

[7] Hoda A. ElMaraghy. 1987. "Artificial intelligence and Robotic Assembly", Springer-Verlag New York Inc.

[8] Cristina Turcu, Cornel Turcu and Vasile Gaitan. 2012. "Integrating robots into the Internet of Things", Issue 6, Volume 6, 2012 430 INTERNATIONAL JOURNAL OF CIRCUITS, SYSTEMS AND SIGNAL PROCESSING

[9] Nils J. Nilsson. 2005. "Human-Level Artificial Intelligence? Be serious!", American Association for Artificial Intelligence, 25th anniversary issue.

[10] Dr. Antoni Diller "HOW EMPIRICISM DISTORTS AI AND ROBOTICS".

[11] Stuart J. Russell and Peter Norvig. 2010. "Artificial Intelligence-A Modern Approach". Prentice-Hall, Inc.

[12] Website: http://www.webopedia.com/TERM/A/artificial_intelligence.html